\documentclass[lettersize,journal]{IEEEtran}
\usepackage{amsmath,amsfonts}
\usepackage{algorithmic}
\usepackage{algorithm}
\usepackage{array}
\usepackage[caption=false,font=normalsize,labelfont=sf,textfont=sf]{subfig}
\usepackage{textcomp}
\usepackage{stfloats}
\usepackage{url}
\usepackage{verbatim}
\usepackage{graphicx}
\usepackage{cite}
\usepackage{caption}
\usepackage{bbding}
\usepackage{pifont}

\usepackage{hyperref}
\usepackage{booktabs}
\usepackage{subfig}
\usepackage{makecell}

\let\oldtwocolumn\twocolumn
\renewcommand\twocolumn[1][]{%
    \oldtwocolumn[{#1}{
    \begin{center}
           \includegraphics[width=0.95\textwidth]{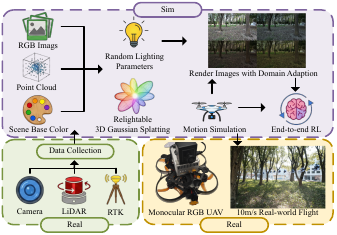}
           \captionof{figure}{The Real-Sim-Real pipeline. By training in a photorealistic 3DGS simulator with active Relightable 3D Gaussian Splatting, our method enables zero-shot, high-speed monocular navigation in complex real-world environments.}
           \label{fig:introduction}
        \end{center}
    }]
}

\begin{document}

\title{Zero-Shot UAV Navigation in Forests via \\ Relightable 3D Gaussian Splatting}

\author{
Zinan~Lv,
Yeqian~Qian*,~\IEEEmembership{Member,~IEEE,}
Chen Sang,
Hao Liu,
Danping~Zou,~\IEEEmembership{Member,~IEEE,}
and~Ming~Yang*,~\IEEEmembership{Member,~IEEE,}

\thanks{Zinan Lv, Yeqiang Qian, Chen Sang, Hao Liu, Danping Zou, and Ming Yang are with School of Automation and Intelligent Sensing, Shanghai Jiao Tong University, Shanghai 200240, China.}
\thanks{Corresponding author: Yeqiang Qian (qianyeqiang@sjtu.edu.cn) and Ming Yang (mingyang@sjtu.edu.cn).}
}

\maketitle

\begin{abstract}
UAV navigation in unstructured outdoor environments using passive monocular vision is hindered by the substantial visual domain gap between simulation and reality. While 3D Gaussian Splatting enables photorealistic scene reconstruction from real-world data, existing methods inherently couple static lighting with geometry, severely limiting policy generalization to dynamic real-world illumination. In this paper, we propose a novel end-to-end reinforcement learning framework designed for effective zero-shot transfer to unstructured outdoors. Within a high-fidelity simulation grounded in real-world data, our policy is trained to map raw monocular RGB observations directly to continuous control commands. To overcome photometric limitations, we introduce Relightable 3D Gaussian Splatting, which decomposes scene components to enable explicit, physically grounded editing of environmental lighting within the neural representation. By augmenting training with diverse synthesized lighting conditions ranging from strong directional sunlight to diffuse overcast skies, we compel the policy to learn robust, illumination-invariant visual features. Extensive real-world experiments demonstrate that a lightweight quadrotor achieves robust, collision-free navigation in complex forest environments at speeds up to 10 m/s, exhibiting significant resilience to drastic lighting variations without fine-tuning.
\end{abstract}



\section{Introduction}
\label{sec:introduction}

\IEEEPARstart{A}{utonomous} navigation of Unmanned Aerial Vehicles (UAVs) in unstructured outdoor environments is crucial, enabling diverse tasks ranging from disaster response~\cite{khan2022emerging, nedjati2016post} to infrastructure inspection~\cite{lu2015uav, trasvina2017unmanned}. To ensure flight safety in these complex settings, contemporary navigation systems predominantly rely on active geometric perception, employing sensors such as LiDAR~\cite{xu2025flying} and depth cameras~\cite{yu2024mavrl, xu2025navrl}. While providing reliable depth information, these active solutions impose severe system-level penalties: they increase payload weight, which degrades flight agility, induce computational latency that bottlenecks high-speed control, and suffer from sensing failures and infrared interference in outdoor sunlight. These constraints motivate the critical inquiry: Is it possible to replicate bird-like, high-speed agility in dense, unstructured forests using solely a passive monocular RGB camera?

Monocular vision-based navigation has emerged as a compelling alternative, offering a pathway toward agile and low-cost autonomy. Learning-based frameworks have demonstrated remarkable success in structured environments, such as drone racing circuits~\cite{kaufmann2023champion, song2023reaching} and indoor corridors~\cite{singla2019memory}, where simulation modeling is straightforward. However, extending these capabilities to unstructured, cluttered outdoor environments remains a complex challenge due to the difficulty of modeling irregular geometries with traditional simulators~\cite{shah2017airsim, song2021flightmare}. Recently, 3D Gaussian Splatting (3DGS)~\cite{3dgs} has gained significant attention in navigation research owing to its photorealistic reconstruction. Approaches like Splat-Nav~\cite{chen2025splat} and \cite{huang2025flying} leverage 3DGS effectively but have predominantly focused on structured indoor environments with stable lighting. Consequently, the distinct challenges of unstructured outdoor settings, particularly characterized by uncontrollable illumination, remain less explored. Addressing this challenge necessitates the capability to simulate dynamic illumination. However, standard relighting approaches often struggle to accurately disentangle geometry from illumination in complex wild scenes, erroneously encoding shadows into the surface albedo. This physical ambiguity limits the synthesis of diverse training data, thereby increasing the risk of overfitting to static capture conditions rather than generalizing to real-world dynamic illumination.

In this paper, we present a novel end-to-end deep reinforcement learning (RL) framework for autonomous UAV navigation, specifically engineered to achieve zero-shot transfer from simulation to the physical domain. By directly mapping raw monocular RGB imagery to continuous control commands, this end-to-end architecture effectively mitigates compounding errors inherent in traditional modular pipelines. Complementarily, the utilization of RL facilitates the acquisition of robust navigation policies for complex behaviors. To enable high-fidelity training, our proposed Real-Sim-Real pipeline, as illustrated in Fig.~\ref{fig:introduction}, leverages unstructured, in-the-wild video sequences to reconstruct photorealistic and geometrically consistent digital twins. Grounding the simulation in empirical real-world data significantly narrows the visual sim-to-real domain gap, enabling agile autonomous navigation in complex environments at speeds exceeding 10 m/s. Remarkably, this agile autonomous flight is achieved using a monocular RGB camera solely without requiring any real-world fine-tuning.

Furthermore, to ensure robustness against the unpredictability of outdoor environments, we introduce Relightable 3D Gaussian Splatting. Addressing the intrinsic limitation of standard 3DGS, where lighting conditions are statically coupled with scene geometry, our method decomposes the scene components to allow for explicit editing of environmental lighting. This capability enables us to augment the training process with a diverse spectrum of synthesized lighting conditions, ranging from strong directional sunlight to diffuse overcast scenarios. By exposing the agent to these variations during simulation, we force the policy to learn illumination-invariant visual features, preventing overfitting to the specific conditions of the data collection time and ensuring reliable performance across different times of day and weather conditions.

The main contributions of this work are summarized as follows:

\begin{itemize}
    \item We develop a novel end-to-end deep reinforcement learning framework for vision-based UAV navigation, which directly maps raw visual observations to continuous control commands, enabling agile autonomous flight without relying on handcrafted features or modular components.
    
    \item To enable effective zero-shot transfer to unstructured outdoor environments, we leverage Relightable 3D Gaussian Splatting to construct high-fidelity simulations with diverse lighting synthesis, ensuring policy robustness to varying real-world illumination conditions.
    
    \item We empirically validate our system through extensive flight experiments, demonstrating that our policy achieves robust, collision-free autonomous navigation at speeds up to 10 m/s in complex forest environments under varying real-world illumination conditions.
\end{itemize}

\section{Related Works}
\label{sec:relatedworks}

\subsection{Vision-Based Autonomous Navigation for UAVs}

Traditional frameworks~\cite{ren2025safety, zhou2020ego, ren2022bubble} typically adopt a modular pipeline, integrating explicit mapping (e.g., voxel grids, ESDFs), localization, and trajectory planning. Representative systems like FASTER~\cite{tordesillas2021faster} ensure safety by maintaining backup trajectories in known free space. While theoretically robust in static settings, these methods require precise state estimation and low-latency depth perception. Maintaining high-resolution dense maps on onboard hardware is computationally expensive. Furthermore, in texture-less, dynamic, or highly cluttered environments where state estimation drifts or depth sensors fail, modular pipelines often become brittle due to error propagation, where a failure in perception leads to catastrophic failure in planning.

To bypass these bottlenecks, learning-based approaches \cite{zhang2025learning, xing2024contrastive, kim2022towards, song2022learning, kulkarni2024reinforcement} have garnered significant attention. Imitation Learning (IL) \cite{hussein2017imitation} has been successfully applied to drone racing~\cite{kaufmann2023champion} and trail following~\cite{giusti2015machine}, yet it suffers from covariate shift. Deep Reinforcement Learning (DRL) offers a robust alternative by allowing agents to learn from failures. Recent works have achieved high-speed flight using active sensors~\cite{xu2025flying, yu2024mavrl, hu2025seeing, zhao2024learning}. However, active sensors impose strict size, weight, and power constraints, thereby severely limiting their practical deployment on agile, sub-250g micro-UAVs.

Recently, 3D Gaussian Splatting has gained significant attention in navigation research owing to its rapid rendering capabilities. Splat-Nav~\cite{chen2025splat} attempts to integrate 3DGS directly into the online planning stack; however, the computational demands of real-time map densification remain prohibitive for resource-constrained micro-UAVs. Similarly, GRaD-Nav++~\cite{chen2025grad} exploits the differentiable nature of 3DGS for trajectory optimization, yet the heavy reliance on iterative rendering and gradient computation poses severe challenges for efficient onboard deployment. Alternatively, \cite{huang2025flying} leverages 3DGS for offline Sim-to-Real training to mitigate these computational constraints. Yet, these methods have primarily been validated in structured indoor environments, leaving the distinct challenges of complex, unstructured outdoor settings, such as dense forests characterized by irregular occlusion and uncontrollable illumination, largely unaddressed.

Monocular vision offers a compelling, lightweight alternative but faces greater perception challenges due to scale ambiguity. Our work targets this specific gap by leveraging the real-time rendering capability of 3DGS to construct a massive, diverse simulation environment. Unlike prior works that either demand heavy onboard mapping or suffer from slow data generation, we utilize 3DGS solely for efficient Sim-to-Real training, allowing the UAV to learn robust depth cues from single RGB images without explicit depth supervision or heavy online computation.

\subsection{High-Fidelity Simulation Environments}

The development of autonomous UAV algorithms faces significant hurdles regarding safety, cost, and scalability when relying exclusively on real-world experimentation. Simulation provides a critical alternative, evolving through distinct technical paradigms \cite{dimmig2024survey} to address the growing demand for fidelity.

Initial simulation platforms prioritized accurate flight dynamics and control systems~\cite{furrer2016rotors, song2021flightmare, todorov2012mujoco}, often integrating seamlessly with standard autopilot stacks like PX4~\cite{PX4GazeboSimulation2024} or ArduPilot~\cite{ArduPilotGazeboSITL2024}. While indispensable for low-level control verification and swarm dynamics, these platforms typically rely on rudimentary geometric primitives and synthetic textures. To address the resulting visual realism gap, researchers increasingly adopted modern game engines (e.g., Unreal Engine, Unity) to render complex scenes~\cite{NVIDIAIsaacSim2024, kulkarni2023aerial}. Prominent frameworks such as AirSim~\cite{shah2017airsim} and AirSim360~\cite{ge2025airsim360} provide rich, visually coherent synthetic data tailored for robotic learning. However, these environments depend on artist-crafted assets or procedural generation. Consequently, they inevitably exhibit stylistic discrepancies, where textures appear artificially clean or repetitive compared to the stochastic complexity of the real world. This discrepancy results in a persistent domain gap, where policies trained on synthetic assets fail to generalize to the intricate, disordered geometry found in real-world unstructured environments.

Recently, data-driven 3D reconstruction has emerged as a promising solution for creating realistic digital twins. Neural Radiance Fields (NeRF)~\cite{mildenhall2021nerf} demonstrated implicit scene modeling capabilities, inspiring works like \cite{adamkiewicz2022vision} to use NeRFs for trajectory optimization. However, the implicit nature of NeRF necessitates computationally expensive volumetric querying for every pixel, rendering it unsuitable for the high-frequency rendering required in large-scale Reinforcement Learning (RL) pipelines. In contrast, 3D Gaussian Splatting (3DGS)~\cite{3dgs} has offered a transformative representation, combining the visual fidelity of NeRFs with the rasterization speed of game engines. Recent robotics applications, such as RAD~\cite{gao2025rad} for autonomous driving and Splat-Nav~\cite{chen2025splat} for dense mapping, have validated its potential. Despite these advances, standard 3DGS-based simulators often inherently entangle static lighting conditions within the scene representation. This lack of environmental controllability restricts the diversity of training data, causing policies to overfit to specific lighting conditions. In this work, we address this limitation by integrating a 3DGS-based simulator with a novel Relightable 3D Gaussian Splatting specifically designed for scalable, randomized RL training.

\subsection{Photometric Domain Adaptation and Relighting}

Traditional Domain Adaptation (DA) methods rely on 2D image-space augmentations. Domain Randomization (DR)~\cite{tobin1970limiting} applies random textures and colors to simulation meshes to force the network to learn geometric invariants. However, this often results in non-physical visual inputs that may hinder the learning of realistic features. Other approaches utilize GANs~\cite{bousmalis2017unsupervised} or Style Transfer~\cite{jackson2019style} to map synthetic images to the real domain. While effective for static images, these generative methods often suffer from temporal inconsistency (flickering) and can generate spurious artifacts, which are detrimental to high-speed control policies that rely on consistent optical flow cues.

In contrast, the advent of 3DGS has enabled more physically grounded appearance editing. Recent approaches like Relightable 3D Gaussians~\cite{gao2024relightable} and GI-GS~\cite{chen2024gi} decompose the scene into geometry, material properties (albedo, roughness), and lighting, allowing for realistic relighting and shadow synthesis. Others, such as GaussCtrl~\cite{wu2024gaussctrl}, leverage diffusion models for semantic editing. However, these rigorous inverse-rendering methods typically incur high computational overheads or require complex deferred shading pipelines. For instance, accurately solving the inverse rendering equation for every frame is computationally prohibitive when running hundreds of parallel environments. Our approach introduces an efficient, geometry-aware illumination editing mechanism that balances physical plausibility with the rendering speed necessary for large-scale policy optimization, enabling effective zero-shot transfer.

\section{Relightable 3D Gaussian Splatting}
\label{sec:relightable_3dgs}

To achieve robust zero-shot transfer, we require a simulation environment capable of synthesizing diverse lighting conditions. However, standard 3DGS inherently couples static environmental illumination with scene geometry, limiting its adaptability. In this section, we first review the standard 3DGS formulation to identify its limitations, and then present our Relightable 3D Gaussian Splatting framework, which explicitly decomposes the scene into geometry, material albedo, and environmental lighting.

\subsection{Preliminaries on Standard 3DGS}
\label{sec:preliminaries}

3D Gaussian Splatting represents a scene as a collection of anisotropic 3D Gaussians. Each Gaussian $G_i$ is defined by a center position $\boldsymbol{\mu}_i \in \mathbb{R}^3$, a covariance matrix $\boldsymbol{\Sigma}_i \in \mathbb{R}^{3\times3}$, an opacity scalar $\alpha_i \in [0,1]$, and view-dependent color coefficients $\mathbf{f}_i$ represented via Spherical Harmonics (SH). The geometry of the $i$-th Gaussian is described by:
\begin{equation}
    G_i(\mathbf{x}) = \exp\left(-\frac{1}{2}(\mathbf{x}-\boldsymbol{\mu}_i)^\top \boldsymbol{\Sigma}_i^{-1} (\mathbf{x}-\boldsymbol{\mu}_i)\right).
\end{equation}
To ensure the covariance matrix $\boldsymbol{\Sigma}_i$ remains positive semi-definite during optimization, it is decomposed into a rotation matrix $\mathbf{R}_i$ and a scaling matrix $\mathbf{S}_i$, such that $\boldsymbol{\Sigma}_i = \mathbf{R}_i \mathbf{S}_i \mathbf{S}_i^\top \mathbf{R}_i^\top$.

During rendering, the 3D Gaussians are projected onto the 2D image plane. The visible color $C(\mathbf{p})$ of a pixel $\mathbf{x}$ is computed using point-based $\alpha$-blending, sorting the $N$ overlapping Gaussians by depth:
\begin{equation}
\label{eq:standard_render}
    C(\mathbf{x}) = \sum_{i \in N} c_i \alpha_i \prod_{j=1}^{i-1} (1 - \alpha_j),
\end{equation}
where $c_i$ is the view-dependent color computed directly from the SH coefficients $\mathbf{f}_i$ based on the viewing direction.

In the standard formulation, the term $c_i$ represents the total observed radiance, which mathematically entangles the intrinsic surface albedo with the specific lighting conditions present at the time of capture. This entanglement prevents the independent modification of illumination, making standard 3DGS unsuitable for synthesizing the novel lighting scenarios required for domain adaptation.

\subsection{Relightable Formulation}

To overcome the static nature of standard 3DGS, where lighting is intrinsically entangled with surface properties, we propose a physically grounded decomposition of the radiance term $c_i$. We re-formulate the rendering process to explicitly model the interaction between learnable diffuse albedo, environmental lighting, and geometric occlusion. The final color $\mathbf{c}_i$ of the $i^{th}$ Gaussian is synthesized via a shading equation that modulates the global environmental light by local visibility and surface material properties:
\begin{equation}
\label{eq:relight_render}
    \mathbf{c}_i = \boldsymbol{\rho}_i \odot \left( \sum_{m=1}^{(l+1)^2} \mathbf{L}_{env}^m \cdot \mathbf{O}_{i}^m \cdot \mathbf{d}_i^m \right),
\end{equation}
where $\boldsymbol{\rho}_i \in \mathbb{R}^3$ represents the learnable diffuse albedo, describing the intrinsic material color independent of illumination. $\mathbf{L}_{env}^m$ denotes the $m^{th}$ SH coefficient of the global environmental lighting. In contrast to standard 3DGS, which optimizes lighting parameters per-Gaussian, we model $\mathbf{L}_{env}$ as a global variable shared across the scene to ensure physical consistency. $\mathbf{O}_{i}^m$ encodes the occlusion coefficient at the Gaussian's center, characterizing the directional light blockage derived from the scene geometry, while $\mathbf{d}_i^m$ is a learnable transfer coefficient that models local geometric interactions, such as surface normal orientation and inter-reflections.

This formulation effectively decouples geometry from illumination. By treating $\mathbf{L}_{env}$ as an independent global variable, we can alter the scene's appearance solely by modifying these coefficients while keeping the geometry $\mathbf{O}_i$ and material $\boldsymbol{\rho}_i$ fixed, which is the cornerstone of our zero-shot domain transfer capability.

\subsection{Occlusion-Aware Visibility Modeling}

Accurate shadow synthesis is critical for realistic outdoor simulation but is often neglected in standard splatting frameworks. To address this, we introduce a pre-computed Occlusion Field to model global light transport via a voxelized probe network.

\noindent \textbf{Visibility Field Construction.} The scene space is discretized into a uniform voxel grid. At each grid node $\mathbf{q}_k$, we render six orthographic depth maps $D_{k,f}$ along the cardinal axes ($f \in \{\pm X, \pm Y, \pm Z\}$) to capture the surrounding geometry. To mitigate geometric noise inherent in 3DGS reconstructions, we apply a robust depth threshold $d_{thresh}$. The binary visibility $V_{k,f}(\mathbf{x})$ for a direction mapped to pixel $\mathbf{x}$ is determined by:
\begin{equation}
    V_{k,f}(\mathbf{x}) = 
    \begin{cases} 
    0, & D_{k,f}(\mathbf{x}) < d_{thresh} \quad (\text{Occluded}) \\ 
    1, & D_{k,f}(\mathbf{x}) \ge d_{thresh} \quad (\text{Visible})
    \end{cases}.
\end{equation}

\noindent \textbf{SH Projection and Interpolation.} To enable efficient differentiable rendering, we project this discrete visibility map onto a Spherical Harmonics basis. The occlusion SH coefficients $\mathbf{B}_{k}$ for probe $k$ are computed by integrating the visibility over the sphere:
\begin{equation}
    \mathbf{B}_{k}^{lm} = \sum_{f} \sum_{\mathbf{x}} V_{k,f}(\mathbf{x}) \cdot Y_{lm}(\omega_{\mathbf{x}}) \cdot \Delta \Omega,
\end{equation}
where $Y_{lm}$ are the real SH basis functions, $\omega_{\mathbf{x}}$ denotes the normalized direction vector corresponding to pixel $\mathbf{x}$, and $\Delta \Omega$ is the solid angle. Finally, to assign occlusion values to the continuous space of Gaussian centers, we perform trilinear interpolation. For an arbitrary Gaussian at position $\boldsymbol{\mu}_i$, its occlusion coefficient $\mathbf{O}_{i}$ is derived from the eight nearest voxel probes $\mathcal{N}(\boldsymbol{\mu}_i)$:
\begin{equation}
    \mathbf{O}_{i} = \sum_{k \in \mathcal{N}(\boldsymbol{\mu}_i)} w_k \cdot \beta_{k} \cdot \mathbf{B}_{k},
\end{equation}
where $w_k$ is the spatial interpolation weight and $\beta_{k}$ is a back-face culling mask ensuring the probe direction aligns with the Gaussian's normal.

\subsection{High-Fidelity HDR Lighting Prior}
\label{sec:hdr_prior}

Jointly optimizing geometry and lighting often leads to ambiguity, where dark textures are incorrectly modeled as shadows or shadows are baked into albedo. To enforce stable disentanglement, we employ a Deep Panorama Lighting (DPL) pipeline to recover high-dynamic-range (HDR) illumination from the Low Dynamic Range (LDR) training images.

Standard LDR images often clip high-intensity sunlight, leading to inaccurate lighting estimation. We first lift the LDR input to an HDR representation using a pre-trained auto-encoder. The global environmental lighting $I_{env}$ is then represented by projecting the HDR map onto SH coefficients $\mathbf{L}_{lm}$:
\begin{equation}
    \mathbf{L}_{lm} = \frac{4\pi}{N_s} \sum_{\omega} I_{HDR}(\omega) Y_{lm}(\omega),
\end{equation}
where $N_s$ is the number of sampling points on the panoramic sphere. 

Crucially, rather than learning these coefficients online, we pre-calculate $\mathbf{L}_{lm}$ for all training frames and freeze them as a strong prior during the 3DGS optimization. This constraint forces the network to attribute residual photometric errors to surface albedo and occlusion updates, ensuring geometrically consistent relighting and enabling the synthesis of realistic shadows under novel illumination conditions.

\begin{figure*}
  \centering
  \includegraphics[width=\textwidth]{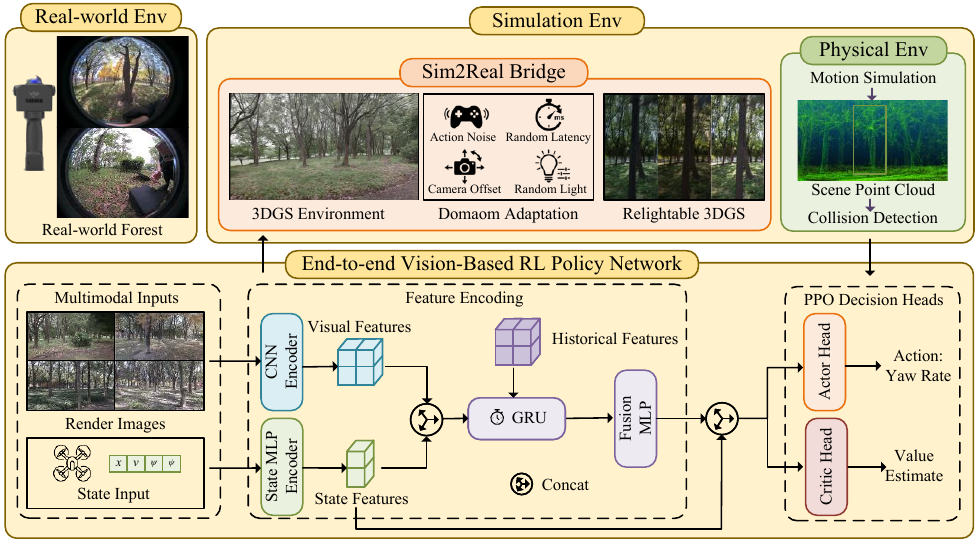}
  \caption{The pipeline of our proposed framework for monocular RGB vision-based autonomous UAV navigation comprises three key stages: 1) Photorealistic Environment Construction: Real-world unstructured scenes are captured and reconstructed using 3D Gaussian Splatting to build a high-fidelity simulator. 2) Sim-to-Real Adaptation: Domain adaptation techniques, including action noise injection, latency simulation, camera pose perturbation, and Relightable 3D Gaussian Splatting, are employed to bridge the visual and dynamics gaps. 3) End-to-end Vision-Based Policy Learning: A reinforcement learning policy processes monocular RGB images and drone state information through CNN and MLP encoders with GRU-based temporal modeling, generating control commands via actor-critic network heads.}
  \label{fig:network}
\end{figure*}

\section{End-to-End Navigation Framework}
\label{sec:method}

As illustrated in Fig.~\ref{fig:network}, we propose a closed-loop framework for zero-shot UAV navigation that seamlessly integrates real-world data acquisition with physical deployment. Our pipeline commences with the reconstruction of unstructured outdoor environments using the proposed Relightable 3DGS to establish a photorealistic and physically consistent simulation substrate. Within this high-fidelity digital twin, we train an end-to-end reinforcement learning policy that maps monocular RGB inputs and proprioceptive states directly to continuous control commands. To bridge the sim-to-real gap, we leverage the disentangled nature of our scene representation to execute aggressive photometric domain adaptation. This strategy exposes the agent to diverse synthesized illumination conditions during training, ensuring that the learned policy remains robust to dynamic lighting and enabling effective zero-shot transfer to challenging real-world forests.

\subsection{Photorealistic Simulation Environment}
\label{sec:method_sim}

To mitigate the visual domain gap, we construct a digital twin of unstructured outdoor environments utilizing the Relightable 3DGS model. Unlike traditional game engines relying on handcrafted assets, our environment faithfully captures the intricate textures and irregular geometries characteristic of real-world forests.

\noindent \textbf{Visual Rendering.} The simulator synthesizes high-fidelity RGB observations $I_t^{sim}$ by querying the Relightable 3DGS model. At each time step $t$, given the drone's pose $\mathbf{T}_t$, the system utilizes the relightable rendering equation to generate views. This mechanism allows for the real-time synthesis of photorealistic imagery that reflects the current, randomized lighting configuration.

\noindent \textbf{Scene Interaction and Collision Detection.} To enable physical interaction, a 3D point cloud $\mathcal{P}$ is extracted from the reconstructed scene by sampling the centers of the Gaussian ellipsoids. A KDTree spatial index is constructed on $\mathcal{P}$ to facilitate efficient proximity queries. The drone's collision volume is approximated as a cylinder with radius $r_{col}$ and height tolerance $h_{tol}$. At each simulation step, we query all scene points within a radius $r_{query} = r_{col} + \delta_{safe}$. A collision event is triggered if any point $\mathbf{p}_{point} \in \mathcal{P}$ satisfies:
\begin{equation}
    \|\mathbf{p}_{d}^{xy} - \mathbf{p}_{point}^{xy}\|_2 \leq r_{col} \quad \text{and} \quad |p_{d}^z - p_{point}^z| \leq h_{tol},
\end{equation}
where $\mathbf{p}_{d}$ denotes the drone's position and $\mathbf{p}_{point}$ represents the position of a scene point.

\noindent \textbf{Quadrotor Dynamics Model.} We employ a simplified yet effective decoupled dynamics model to balance physical fidelity with computational efficiency. The drone's kinematic state is defined as $\mathbf{x} = [\mathbf{p}^\top, \mathbf{v}^\top, \psi, \dot{\psi}]^\top$. To ensure numerical stability, the dynamics are integrated using $\mathcal{N}$ substeps within each control interval $\Delta t$. Within each substep, the yaw dynamics follow a first-order response:
\begin{equation}
    \ddot{\psi} = K_\psi (u - \dot{\psi}),
\end{equation}
where $K_\psi$ is the yaw control gain and $u$ is the target yaw rate. The total force acting on the drone is computed as:
\begin{equation}
    \mathbf{F}_{\text{total}} = \mathbf{F}_{\text{thrust}} + \mathbf{F}_{\text{drag}} + \mathbf{F}_{\text{gravity}} + \mathbf{F}_{\text{altitude}},
\end{equation}
where $\mathbf{F}_{\text{thrust}}$ is the thrust force aligned with the current heading direction $\mathbf{d}=[\cos\psi, \sin\psi, 0]^\top$. The aerodynamic drag is modeled as $\mathbf{F}_{\text{drag}} = -c_d \|\mathbf{v}\| \mathbf{v}$ with drag coefficient $c_d$, and $\mathbf{F}_{\text{altitude}}$ is the output of a PID controller maintaining a constant flight altitude. The resultant acceleration is used to update the velocity $\mathbf{v}$ and position $\mathbf{p}$ via semi-implicit Euler integration.

\begin{figure}[!tb]
\centering
\includegraphics[width=\linewidth]{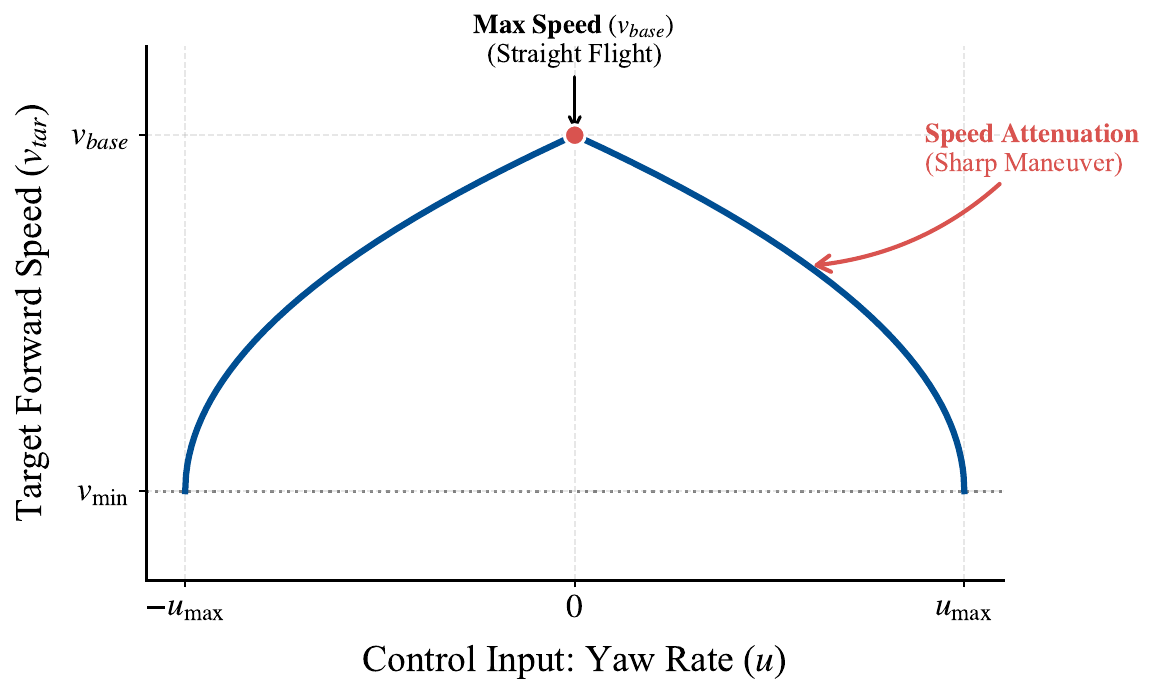}
\caption{Visual illustration of the adaptive speed schedule defined in Eq.~\ref{eq:speed}. The target forward speed $v_{tar}$ is maximized at $v_{base}$ during straight flight and is smoothly attenuated towards $v_{\min}$ as the yaw rate $|u|$ approaches the limit $u_{\max}$, preventing sideslip during sharp turns.}
\label{fig:speed}
\end{figure}

\noindent \textbf{Adaptive Speed Schedule.} To ensure dynamic stability during high-speed maneuvers, we incorporate a forward dynamics schedule that couples the target forward velocity $v_{tar}$ with the yaw rate action $u$. The velocity is modulated as:
\begin{equation}
\label{eq:speed}
    v_{tar} = v_{\min} + (v_{base} - v_{\min}) \cdot \left(1 - \frac{|u|}{u_{\max}}\right)^{0.5},
\end{equation}
where $v_{base}$ and $v_{\min}$ represent the nominal and minimum forward speeds, respectively. As visualized in Fig.~\ref{fig:speed}, this profile enforces a smooth attenuation of forward velocity as the yaw rate increases, thereby preventing sideslip during aggressive turns.

\subsection{End-to-End Vision-Based RL Policy Network}
\label{sec:method_rl}

\noindent \textbf{Network Architecture.} We develop a neural network-based policy designed to map multimodal observations to continuous control commands. At each time step $t$, the policy processes two distinct input streams: a high-dimensional visual observation $I_t \in \mathbb{R}^{H \times W \times 3}$ and the drone's kinematic state vector $\mathbf{s}_t$. The state is defined in relative coordinates as $\mathbf{s}_t = [\mathbf{p}_{rel}^\top, \mathbf{v}^\top, \psi, \dot{\psi}]^\top$. The network outputs a continuous scalar action $a_t \in [-u_{\max}, u_{\max}]$, corresponding to the target yaw rate.

The architecture integrates three specialized modules. A Convolutional Neural Network (CNN) serves as the visual backbone, extracting spatial geometric features from $I_t$. Concurrently, a Multi-Layer Perceptron (MLP) encodes the proprioceptive state $\mathbf{s}_t$ into a compact latent embedding. To address the partial observability inherent in monocular navigation, we incorporate a Gated Recurrent Unit (GRU)~\cite{chung2014empirical} to capture sequential dynamics. The GRU maintains a hidden state by processing the concatenated visual and state features, effectively aggregating temporal context. The final control output is generated via a fusion module followed by fully connected layers that parameterize a Gaussian distribution $\pi(a_t | \cdot)$ for action sampling.

\noindent \textbf{Reward Function Design.} To guide the policy toward safe and efficient navigation, we design a composite reward function $r_t$ consisting of dense shaping signals. The formulation and weights are detailed in Table~\ref{tab:rewards}.

\begin{table}[h]
\renewcommand{\arraystretch}{1.5}
\centering
\caption{Components and weights of the reward function}
\label{tab:rewards}
\resizebox{\columnwidth}{!}{
\begin{tabular}{l c c}
\hline
\textbf{Reward Component} & \textbf{Mathematical Formulation} & \textbf{Weight} \\
\hline
Progress Reward & $r_{progress} = d_{t-1} - d_t$ & $1.0$ \\
Alignment Reward & $r_{align} = \cos(\psi_{target} - \psi_t)$ & $0.1$ \\
Obstacle Penalty & $r_{obstacle} = \min\left(0, \frac{d_{obs} - R_{safe}}{R_{safe}}\right)$ & $0.2$ \\
Success Reward & $r_{success} = \mathbb{I}(d_t < R_{goal})$ & $100.0$ \\
Collision Penalty & $r_{collision} = \mathbb{I}(\text{collision})$ & $-50.0$ \\
\hline
\end{tabular}
}
\end{table}

The progress reward $r_{progress}$ incentivizes the agent to approach the goal by maximizing the reduction in Euclidean distance $d_t$. The alignment reward $r_{align}$ promotes smooth path following. A sparse success reward $r_{success}$ is granted upon reaching the target vicinity. To ensure safety, an obstacle penalty $r_{obstacle}$ imposes a negative cost linearly proportional to proximity when the distance to the nearest obstacle falls below a safety margin. Finally, a discrete collision penalty $r_{collision}$ is applied upon impact. The total reward is the weighted sum:
\begin{equation}
\label{eq:reward}
    r_t = \sum_i \lambda_i \cdot r_{t}^i.
\end{equation}

\noindent \textbf{Training Curriculum.} The objective is to optimize the policy parameters $\theta$ to maximize the expected cumulative reward using Proximal Policy Optimization (PPO)~\cite{schulman2017proximal}. The training curriculum is structured into two distinct phases. Initially, the policy is trained on 3DGS scenes with static, original illumination to establish fundamental geometric understanding. Subsequently, the second phase activates the Relightable 3D Gaussian Splatting to introduce randomized lighting variations. This two-stage curriculum facilitates photometric domain adaptation, enabling the policy to learn robust features invariant to lighting changes.

\subsection{Training with Photometric Domain Adaptation}
\label{sec:method_adaptation}

To enable zero-shot transfer to the physical world, we leverage the disentangled structure of our Relightable 3DGS to implement an aggressive Photometric Domain Adaptation strategy.

Instead of training on static lighting, we randomize the global illumination coefficients $\mathbf{L}_{env}$ at the start of every training episode. We apply three types of perturbations to the source HDR lighting coefficients:
\begin{itemize}
    \item \textbf{Rotation:} The SH coefficients are rotated around the vertical axis to simulate varying times of day and solar azimuths.
    \item \textbf{Intensity Scaling:} The coefficients are scaled to mimic diverse exposure levels and cloud cover densities.
    \item \textbf{Chromatic Tinting:} Color shifts are applied to the coefficients to simulate different weather conditions (e.g., warmer tones for dusk, cooler tones for overcast days).
\end{itemize}

By exposing the RL agent to this continuous spectrum of synthesized illumination while strictly preserving the underlying geometry $\mathbf{O}_i$ and material albedo $\boldsymbol{\rho}_i$, we compel the policy to learn robust geometric features that are invariant to photometric variations. This ensures that the navigation policy remains reliable when deployed in real-world forests where lighting conditions are unpredictable and dynamic.

\begin{figure*}[!tb]
\centering
\includegraphics[width=\linewidth]{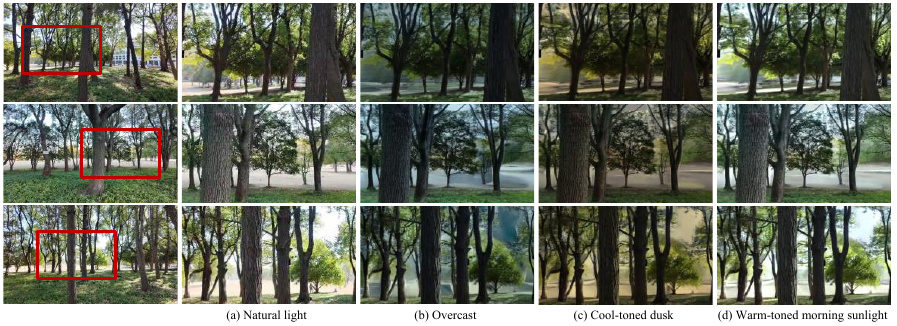}
\caption{Examples of photorealistic Relightable 3D Gaussian Splatting. The columns display the original natural light (a) and synthesized variations: overcast (b), cool-toned dusk (c), and warm-toned morning sunlight (d) across different outdoor scenes.}
\label{fig:light}
\end{figure*}

\section{Experiment}
\subsection{Experiment Set}
\noindent \textbf{High-Fidelity 3DGS Simulation Environment Construction.}
To establish a photorealistic training and testing ground, we captured a diverse set of real-world forest scenes. The data acquisition was performed using a handheld device equipped with a stereo fisheye camera rig and a high-precision LiDAR integrated with a Real-Time Kinematic (RTK) positioning system. This sensor suite enabled the collection of synchronized, geo-referenced multimodal data. The RTK system provided centimeter-accurate global poses, which served as the initial camera parameters for the 3DGS optimization pipeline, while the LiDAR delivered a dense initial point cloud for scene geometry initialization. In total, ten distinct forest scenes were captured, each spanning an area of approximately $60 \times 60$ meters. These scenes were subsequently reconstructed into high-fidelity 3D models using the 3DGS framework, forming the core of our photorealistic simulator.

\noindent \textbf{Simulation Implementation Details.}
The navigation policy is trained using Proximal Policy Optimization. Training is conducted in two phases across parallel simulation environments. The simulation operates at a control frequency of 10 Hz, corresponding to a timestep of $\Delta t = 0.1$ s. The drone's forward speed follows an adaptive schedule up to a maximum of $v_{\max}=10$ m/s, with a maximum yaw rate command of $u_{\max}=1.0$ rad/s. For each training episode, a valid start-goal pair is randomly sampled from the scene; a pair is deemed valid if a collision-free navigation path exists between the points and the Euclidean distance exceeds 30 meters, ensuring episodes of sufficient challenge. The drone's collision volume is modeled as a cylinder with a radius of $r_{col}=0.3$ m and a height tolerance of $h_{tol}=0.2$ m. The safety distance threshold for the obstacle-aware reward component is set to $R_{safe}=2.0$ m. An episode terminates successfully when the drone arrives within 2.0 m of the goal position. All experiments were executed on a workstation equipped with an NVIDIA RTX 3090 GPU (24GB memory).

\begin{table}[!tb]
\caption{Parameter Randomization for Domain Adaptation.}
\begin{center}
\begin{tabular}{l c}
\hline
Parameter & Distribution / Value \\
\hline
Action Noise & $\mathcal{N}(0, 1.0^2)$ \\
Latency Delay & $\mathcal{U}(0, 80)$ ms \\
Control Interval & $\mathcal{U}(10, 100)$ ms \\
XY Position Noise & $\mathcal{N}(0, 0.05^2)$ m \\
Z Position Noise & $\mathcal{N}(0, 0.03^2)$ m \\
Velocity Noise & $\mathcal{N}(0, 0.08^2)$ m/s \\
Camera Position Offset & $\mathcal{U}(-0.1, 0.1)$ m \\
Camera Orientation Offset & $\mathcal{U}(-5^\circ, 5^\circ)$ \\
Relightable 3DGS & Rotation, Intensity, Chromatic Tinging \\
\hline
\end{tabular}
\end{center}
\label{tab:domain_adaptation}
\end{table}

To bridge the reality gap and ensure robust zero-shot transfer, we employ a comprehensive domain randomization strategy that perturbs dynamics, perception, and photometric parameters during training, as detailed in Table~\ref{tab:domain_adaptation}. To mimic the imperfections of physical hardware, we introduce Gaussian noise into state estimates (position, velocity, and height) and simulate actuator dynamics through action noise, variable control intervals, and stochastic latency. Perception robustness is further enforced by applying random perturbations to the camera's extrinsic pose to account for calibration errors. Crucially, addressing the visual domain gap, we leverage our Relightable 3D Gaussian Splatting framework to go beyond standard 2D augmentations. By explicitly manipulating the global spherical harmonic coefficients within the scene representation, we randomize the illumination direction, intensity, and spectral properties, effectively synthesizing diverse and physically consistent outdoor lighting conditions that prepare the agent for unstructured real-world environments.

\noindent \textbf{Real-world Implementation Details.}
For real-world validation, we employed a custom-built lightweight quadrotor platform. The hardware configuration was designed to prioritize low weight and onboard computational capability for autonomous vision-based flight. The core flight controller runs the PX4 open-source autopilot stack. For perception and high-level decision-making, the platform is equipped with an NVIDIA Jetson Orin NX onboard computer. The primary exteroceptive sensor is a monocular RGB camera operating at 60 frames per second. An inertial measurement unit (IMU) integrated with a GNSS receiver provides state estimation; this fused positioning and attitude data are refined in real-time via an extended Kalman filter (EKF) \cite{ribeiro2004kalman} to supply robust odometry. No LiDAR, depth cameras, or other active ranging sensors were used, ensuring the system relies purely on passive monocular vision.

\begin{figure}[!tb]
\centering
\includegraphics[width=\linewidth]{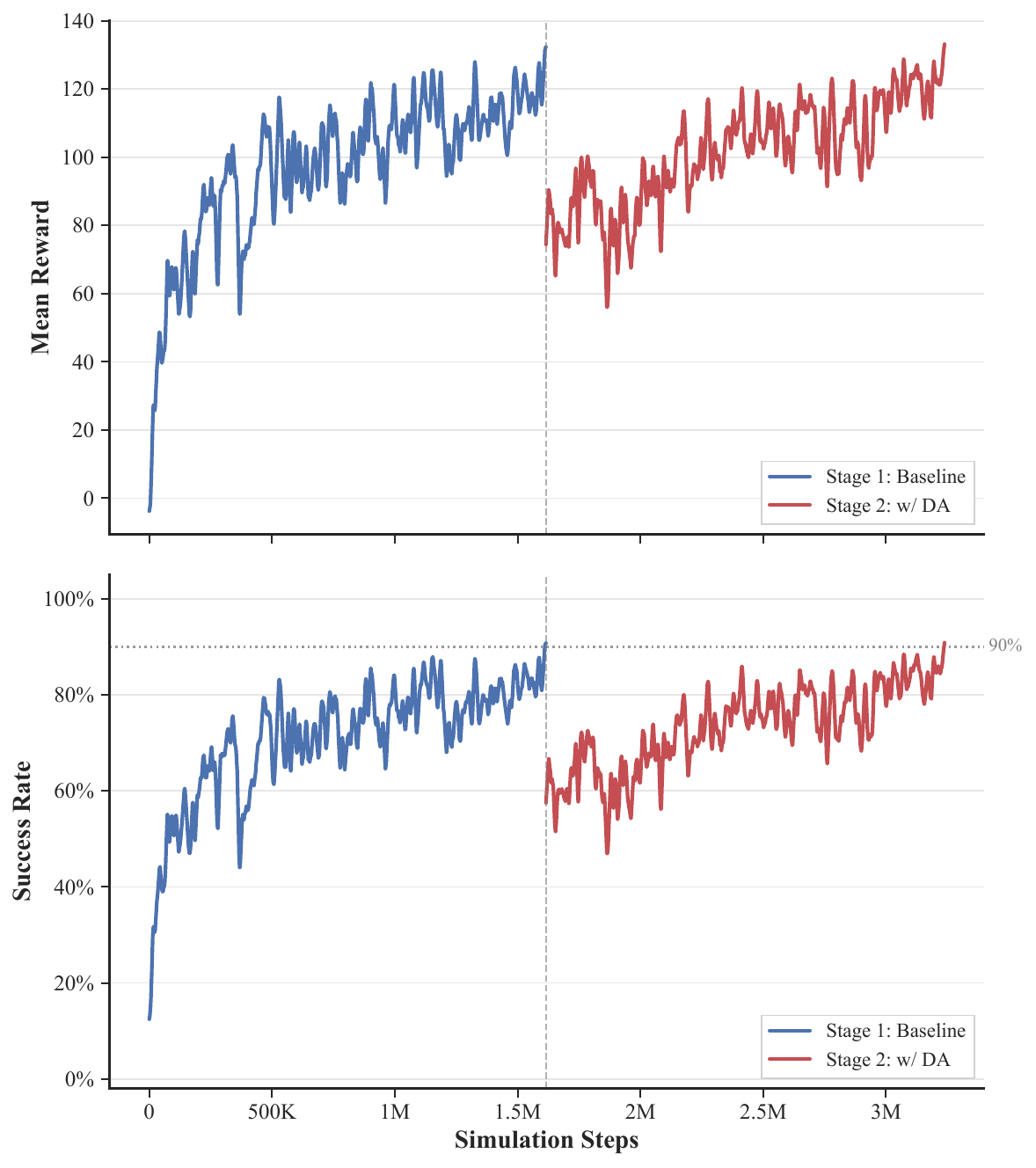}
\caption{Simulation training performance evolution across two stages. The top panel illustrates the mean reward, while the bottom panel displays the navigation success rate over simulation steps. The vertical gray dashed line marks the curriculum transition from Stage 1 (Baseline training, blue curves) to Stage 2 (training with Domain Adaptation, red curves) at approximately 1.6M steps.}
\label{fig:sim-result}
\end{figure}

\subsection{Simulation Results}

\noindent \textbf{Photorealistic Domain Adaptation via Relightable 3D Gaussian Splatting.}
Outdoor environments are characterized by high-frequency photometric variations, presenting a formidable challenge for vision-based navigation. Standard 3DGS reconstructions inherently entangle static environmental illumination with scene geometry, restricting the simulation to the specific lighting conditions captured during data collection. This limitation creates a significant sim-to-real domain gap, as policies trained on static illumination often fail to generalize to the dynamic lighting scenarios encountered in physical deployment.

To systematically bridge this gap, we integrate our Relightable 3D Gaussian Splatting module into the training pipeline. This mechanism allows for the synthesis of diverse, photorealistic illumination conditions while strictly preserving the underlying geometric consistency. As demonstrated in Fig.~\ref{fig:light}, starting from a baseline reconstruction under natural light (a), our approach generates physically plausible variations ranging from diffuse overcast skies (b) to cool-toned dusk (c) and warm-toned morning sunlight (d). By augmenting the training curriculum with this continuous spectrum of synthesized lighting, we compel the navigation policy to learn robust visual representations that are invariant to photometric shifts, thereby significantly enhancing its zero-shot transfer capability.

\begin{figure*}[!tb]
\centering
\includegraphics[width=\linewidth]{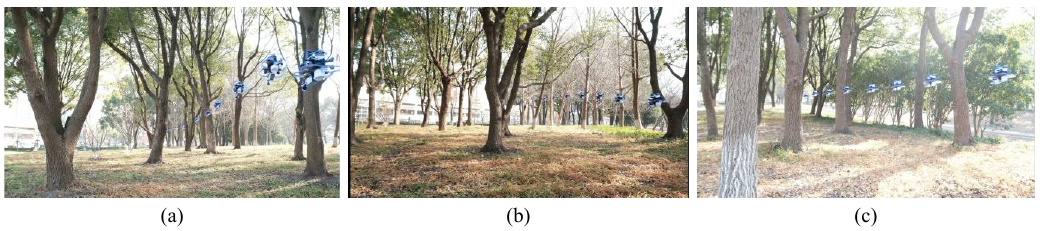}
\caption{Real-world flight trajectories across multiple unstructured forest environments. Each subplot shows a successful navigation trial from a distinct location, with the drone's path overlaid in color. The trajectories demonstrate the policy's ability to generalize to various cluttered scenes and execute collision-free navigation.}
\label{fig:real-world-scene}
\end{figure*}

\begin{figure*}[!tb]
\centering
\includegraphics[width=\linewidth]{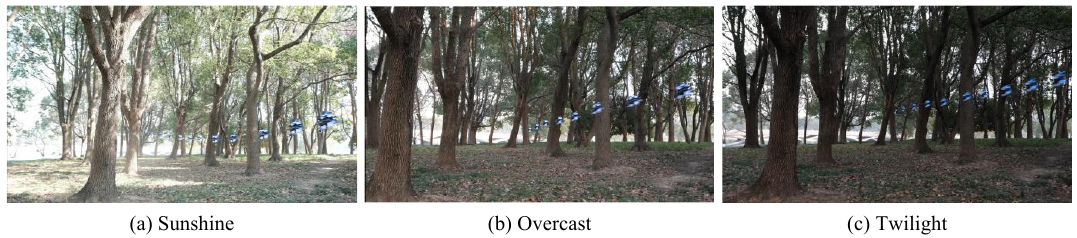}
\caption{Flight trajectory comparison under different natural illumination conditions. Each column represents trials conducted under (a) Sunshine, (b) Overcast, and (c) Twilight lighting in the same forest area. The consistent, goal-directed paths across all three conditions demonstrate the illumination robustness achieved through our domain adaptation method.}
\label{fig:real-world-light}
\end{figure*}

\begin{table*}[!tb]
\caption{Quantitative comparison with RL-based UAV navigation frameworks. We evaluate methods based on sensor modality, input observation, deployment environment, and illumination adaptation capability. The symbol * denotes results reported in the simulation.}
\centering
\begin{tabular}{l c c c c c c}
\hline
Method & Sensor & Observation & Environment & Max Speed & Success Rate & Random Light \\
\hline
FPCRL\cite{xu2025flying} & LiDAR & Point Cloud & Indoor \& Outdoor & 3.0 m/s & 80\% & - \\
MAVRL\cite{yu2024mavrl} & RGB-D Camera & Depth Image & Indoor & 5.5 m/s* & 60\%* & - \\
NavRL\cite{xu2025navrl} & RGB-D Camera & Depth Image & Indoor & 2.0 m/s* & 80\%* & - \\
Reinforcement\cite{kulkarni2024reinforcement} & RGB-D Camera & Depth Image & Indoor & 1.2 m/s & 70\% & - \\
Differentiable\cite{zhang2025learning} & RGB-D Camera & Depth Image & Indoor \& Outdoor & 7-23m/s & 90\% & - \\
Learning\cite{zhao2024learning} & RGB-D Camera & 3D Occupancy & Indoor \& Outdoor & 3.5 m/s & -  & -\\
\hline
D3QN\cite{kim2022towards} & Monocular & Depth Image & Indoor & 0.4 m/s & - & \ding{55} \\
Seeing\cite{hu2025seeing} & Monocular & Optical Flow & Indoor \& Outdoor & 6.0 m/s & 60\% & \ding{55} \\
CAD2RL\cite{sadeghi2016cad2rl} & Monocular & RGB & Indoor & 0.2 m/s* & 40\%* & \ding{55} \\
Flying\cite{huang2025flying} & Monocular & RGB & Indoor & 2.0 m/s & 80\% & \ding{55} \\
Splat-Nav\cite{chen2025splat} & Monocular & RGB & Indoor & 1.5 m/s & 99\% & \ding{55} \\
\hline
Ours & Monocular & RGB & Outdoor & 10.0 m/s & 80\% & \Checkmark \\
\hline
\end{tabular}
\label{tab:com}
\end{table*}

\noindent \textbf{Two-Stage Training and Domain Adaptation Performance.} 
Figure~\ref{fig:sim-result} presents the learning progress of our navigation policy, quantified by the mean reward and the navigation success rate throughout the simulation training process. The training follows a two-stage curriculum, delineated by a vertical dashed line at approximately 1.6M simulation steps.

Stage 1: Baseline Policy Training. In the initial stage (blue curves), the policy is trained in the standard 3DGS simulator without domain adaptation. The results show rapid initial learning, with the mean reward converging to a plateau around 120 and the navigation success rate stabilizing near 90\%. This baseline phase establishes a competent policy for the original, static environment.

Stage 2: Domain Adaptation Integration. In the second stage (red curves), we activate our full suite of Domain Adaptation (DA) techniques, including the proposed Relightable 3D Gaussian Splatting. Introducing these environmental variations initially causes a predictable, transient performance dip as the policy adapts to the increased complexity. Crucially, the policy recovers and then surpasses its previous performance, with the success rate exceeding the 90\% target threshold. The final mean reward also reaches a higher plateau compared to Stage 1. This progression validates our two-stage curriculum: a stable policy is first established in a consistent environment, and then its robustness and generalization are systematically enhanced through diversified training facilitated by domain adaptation.

\subsection{Real-world Flight}

The core objective of our framework is to achieve robust, high-speed autonomous navigation in complex, unstructured outdoor environments using only monocular vision. To this end, we deployed our policy, trained exclusively within the 3DGS simulator with Relightable 3D Gaussian Splatting, onto our custom quadrotor platform for extensive real-world validation.

As illustrated in Fig.~\ref{fig:real-world-scene}, we conducted flight tests across multiple distinct forest environments. Each subfigure depicts a complete navigation trial, overlaying the drone's estimated trajectory onto the scene. The trajectories demonstrate the policy's ability to successfully plan and execute collision-free paths through dense clutter, navigating around trees and through narrow gaps to reach designated target points.

A key claim of our work is the policy's robustness to significant illumination changes. Fig.~\ref{fig:real-world-light} validates this claim by comparing flight trajectories under three challenging lighting conditions: Sunshine, Overcast, and Twilight. The trajectories remain consistently smooth and goal-directed across all conditions. In the bright sunshine condition, the policy successfully manages high-contrast shadows that could be mistaken for obstacles. Under the uniform lighting of overcast skies, it maintains precise navigation. Most notably, in the low-light, high-dynamic-range twilight setting, the policy continues to operate reliably.

To benchmark our system against RL-based navigation frameworks, we present a quantitative comparison in Table~\ref{tab:com}. While prior learning-based methods are predominantly restricted to indoor environments, rely on active depth sensors (e.g., RGB-D, LiDAR), or lack explicit mechanisms for photometric adaptation, our method achieves a breakthrough in unstructured outdoor settings using a passive monocular camera. Distinctively, our policy supports robust autonomous flight at speeds up to 10.0 m/s under varying illumination, significantly outperforming existing baselines and validating the efficacy of our relightable Sim-to-Real pipeline.

\subsection{Ablation Study}

\begin{figure}[!tb]
\begin{center}
\includegraphics[width=0.95\linewidth]{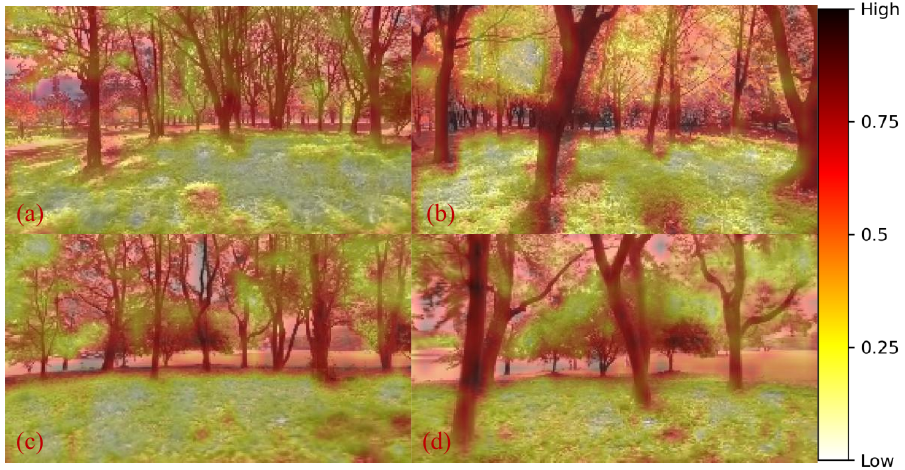}
\caption{Qualitative visualization of learned spatial attention maps overlaid on input RGB frames. Warmer colors (red/orange) indicate regions with higher activation weights within the feature extractor. The examples demonstrate that the policy consistently focuses on salient obstacles critical for collision avoidance.}
\label{fig:attention}
\end{center}
\end{figure}

\noindent \textbf{Visual Feature Learning.}
To elucidate the perceptual strategies acquired by our end-to-end monocular vision policy, we perform an introspective analysis of its convolutional feature extractor. By computing and visualizing spatial attention maps from key intermediate network layers, we identify which regions within the input RGB frame are prioritized during the policy’s decision-making process. These attention heatmaps highlight the image areas that yield the strongest activations, effectively revealing the focus of the model.

As shown in Fig.~\ref{fig:attention}, overlaying the attention maps onto the original images provides a qualitative understanding of the policy’s perceptual priorities. The visualizations consistently demonstrate that the network learns to concentrate on structural features directly relevant to safe navigation: the boundaries of nearby obstacles such as trees and branches, as well as openings that indicate traversable corridors. Importantly, this implicit obstacle awareness emerges purely through reinforcement learning, without any direct geometric or segmentation supervision. This result confirms that our pipeline successfully distills essential navigational cues from raw visual input, validating that the learned visual representations are both meaningful and functionally aligned with the task of high-speed collision avoidance in clutter.

\begin{table}[!tb]
\centering
\caption{Ablation study on Relightable 3D Gaussian Splatting for real-world domain adaptation. The table reports real-world navigation success rates (successful trials / total trials) in unstructured outdoor scenes under three distinct illumination conditions at a target speed of 10 m/s.}
\begin{tabular}{c c c c}
\hline
Relightable 3DGS & Sunshine & Overcast &  Twilight \\
\hline
 & 6/10 & 9/10 & 3/10 \\
\checkmark & \textbf{8/10} & \textbf{10/10} & \textbf{8/10} \\
\hline
\end{tabular}
\label{tab:ablation-ad}
\end{table}

\noindent \textbf{Relightable 3D Gaussian Splatting Domain Adaptation.}
The ultimate test of our domain adaptation strategy is its ability to improve real-world flight robustness under natural illumination changes. We conducted field trials in three distinct outdoor lighting conditions to evaluate our navigation policy, comparing performance with and without Relightable 3D Gaussian Splatting.

As shown in Table \ref{tab:ablation-ad}, we tested the policy navigating unstructured forest terrain at 10 m/s under bright sunshine (with strong directional shadows), overcast skies (with soft diffuse illumination), and dusk (with low ambient illumination). The baseline policy, trained using only the static illumination from the original 3DGS reconstruction, showed clear sensitivity to these variations. Its performance was notably unstable, with a high failure rate in high-contrast sunshine and near-complete failure in the perceptually challenging twilight condition.

In contrast, the policy enhanced with our Relightable 3D Gaussian Splatting domain adaptation demonstrated consistent and superior robustness. Exposure to a synthetically diversified spectrum of lighting conditions during training enabled the policy to learn an illumination-invariant representation of the environment. This is evidenced by sustained high success rates across all tested conditions, with the most significant improvement observed in the challenging dusk scenario. These real-world results confirm that our Relightable 3D Gaussian Splatting method is a critical component for enabling reliable, high-speed vision-based navigation in the face of natural photometric variance.

\section{Conclusion}
\label{sec:conclusion}

In this paper, we present a novel end-to-end framework to address the challenge of high-speed monocular UAV navigation in unstructured outdoor environments. To bridge the critical visual sim-to-real gap, we leverage a photorealistic simulation environment built upon our novel Relightable 3D Gaussian Splatting representation, which allows for controllable and physically consistent illumination synthesis. By training within this high-fidelity environment, the reinforcement learning policy is able to acquire robust visual representations that generalize across lighting changes. Consequently, we enable effective zero-shot transfer to the physical world. Experimental results demonstrate successful, collision-free flight at speeds up to 10 m/s in cluttered forest scenarios under diverse lighting conditions, such as strong sunlight and twilight. Ultimately, our work demonstrates the feasibility of achieving agile and perception-aware autonomy on lightweight aerial platforms in the real world.

\bibliographystyle{unsrt}
\bibliography{mybibfile}

\vfill

\end{document}